**Empirical Analysis of Zipf's Law, Power law, and Lognormal Distributions in Medical Discharge Reports**

Juan C Quiroz PhD, Liliana Laranjo MD MPH PhD, Catalin Tufanaru MD MPH PhD, Ahmet Baki Kocaballi PhD, Dana Rezazadegan PhD, Shlomo Berkovsky PhD, Enrico Coiera MBBS PhD

J.C.Q., L.L., C.T., A.B.K., D.R., S.B., and E.C.: Australian Institute of Health Innovation, Macquarie University, Sydney, Australia

J.C.Q.: Centre for Big Data Research in Health, UNSW, Sydney, Australia

L.L.: University of Sydney, Sydney, Australia

A.B.K.: Faculty of Engineering and IT, University of Technology Sydney

D.R.: Swinburne University of Technology, Department of Computer science and Software Engineering, Melbourne, Australia

Corresponding Author: Juan C Quiroz

Centre for Big Data Research in Health

UNSW, Sydney, Australia

juan.quiroz@unsw.edu.au

**Keywords:** Data mining, MIMIC-III dataset, machine learning, maximum likelihood estimation, power-law with exponential cut-off, statistical distributions.

**ABSTRACT**

Background: Bayesian modelling and statistical text analysis rely on informed probability priors to encourage good solutions.

Objective: This paper empirically analyses whether text in medical discharge reports follow Zipf's law, a commonly assumed statistical property of language where word frequency follows a discrete power-law distribution.

Method: We examined 20,000 medical discharge reports from the MIMIC-III dataset. Methods included splitting the discharge reports into tokens, counting token frequency, fitting power-law distributions to the data, and testing whether alternative distributions—lognormal, exponential, stretched exponential, and truncated power-law—provided superior fits to the data.

Result: Discharge reports are best fit by the truncated power-law and lognormal distributions. Discharge reports appear to be near-Zipfian by having the truncated power-law provide superior fits over a pure power-law.

Conclusion: Our findings suggest that Bayesian modelling and statistical text analysis of discharge report text would benefit from using truncated power-law and lognormal probability priors and non-parametric models that capture power-law behavior.

## INTRODUCTION

Machine learning (ML) is increasingly being used for healthcare applications [1]. Amongst potential applications, analysis of medical free text data provides opportunities for information extraction, representation learning, disease prediction, phenotyping, summarization, and discharge report generation [2–6]. However, analysis of medical text data is challenging due to inconsistent and absent structure, noisy text, complex vocabulary, medical abbreviations, and ambiguity in language [7–9]. The performance of ML models can be improved by first understanding the problem domain and the properties of the data being modelled, which can then guide the selection of appropriate algorithms and parameters. For Bayesian modelling, a probability prior (i.e. uniform, Gaussian, Dirichlet) is used to encode the probabilistic assumptions about a problem domain or the data being modelled [10].

Recognising the statistical properties of medical text can guide the design of ML and Bayesian approaches. A widely used statistical property of language is Zipf's law, which states that word frequency follows a discrete power-law distribution and is inversely proportional to word rank [11]. That is, the most frequent word occurs twice as often as the second most frequent word, three times as often as the third most frequent word, and so on. A few top-ranked words are very frequent, while numerous words with low frequency make up the long *tail* of the distribution. Zipf's law has been shown to hold for various text corpora and natural languages [11–13] and this power-law property has been incorporated into Bayesian language models, topic models, and word embeddings [14–16]. Thus, knowing whether medical discharge reports follow a power-law or alternative distribution would allow for the selection of appropriate probability priors for Bayesian modelling, ML models, and model parameterization. This in turn can benefit ML text applications ranging from automated extraction of information from medical notes to automatic summarization of medical text.

To date, no work has explored Zipf's law in medical and clinical language, such as discharge report text. Zipf's law has been explored in clinical codes [17,18], clinical diagnoses [18], virtual patient physiologic derangements [19], and epidemiology [20], but not in unstructured medical text. We aim to test whether medical discharge reports in the MIMIC-III dataset [21] follow Zipf's law, a power-law distribution, and assess the fit of alternative probability distributions.

## METHODS

### *Data Set*

We used a sample of 20,000 medical discharge reports from the MIMIC-III dataset, comprising information from patients admitted to critical care units at Beth Israel Deaconess Medical Center [21]. The dataset is composed of about 26,365,351 word occurrences and 118,021 unique words. For our primary analysis, we did not apply lemmatization (removing the inflectional endings of words, resulting in a common base word) to preserve as much of the original text as possible and because prior work has shown that Zipf's law applies for words and lemmas [22]. We applied lemmatization as part of sensitivity analysis. See Appendix A for tokenization details.

### *Power-law fitting*

A power-law is a probability distribution with the form:

$$p(x) \propto x^{-\alpha} \qquad for\ x \geq x_{min},$$

where $x_{min}$ indicates the minimum value where the scaling relationship of the power-law begins [23]. We followed the methods described by Clauset et al. [24] for analysing power-law distributed data by estimating the parameters $x_{min}$ and $\alpha$ using maximum likelihood estimation (MLE)—see Appendix B for details. Concluding that the power-law is the best description for the data follows two steps [23–25]: (1) a goodness-of-fit test to determine whether the power-law is an appropriate fit for the data (P>0.10 indicates power-law is plausible, P≤ 0.10 indicates power-law is plausible [24]); and (2) comparing the fits of the power-law with alternative heavy-tailed distributions using a log likelihood ratio test. The alternative distributions included the lognormal, exponential, stretched exponential, and truncated power-law (also referred to as power-law with an exponential cut-off).

A bootstrapping hypothesis test was used for the goodness-of-fit test between the data and the power-law distribution. We used the powerlaw Python package to fit the data using MLE, to compute the likelihood ratio tests, and for plotting the fits to the data [23]. We used the poweRlaw R package for the bootstrapping hypothesis test and for estimating the uncertainty of the model parameters (1000 bootstraps) [25]. For discharge reports and each subsection, we indicate the support for the power-law using the ordinal scale first presented in [24]: “none” (not power-law distributed), “moderate” (power-law is a good fit, but alternatives provide good fits as well), “good” (the power-law is a good fit and the alternatives considered are not good fits), and “with cut-off” (the truncated power-law provides a better fit than the pure power-law).  We also present plots of the complementary cumulative distribution function (CCDF) of the word frequency vs word rank to visualize the word frequency data, the fit of the power-law, and the fit of alternative probability distributions [24].

***Discharge Report Subsections***

We also analysed whether subsections of the discharge reports follow a power-law distribution, focusing on the major delineated subsections: allergies, family history, history of present illness, and social history. These sections were chosen due to them being commonly collected during clinical visits in a wide variety of settings, such as general practitioner (GP) consultations, emergency department visits, and specialist visits.

***Sensitivity Analysis***

Discharge reports are poorly formatted, with various words consisting of a combination of letters, letters, and various punctuation marks. As such, a naïve tokenizing based on white space may not break up words appropriately. To assess the effect of a different type of tokenization, we used the spaCy Python library to tokenize the text (model “en_core_web_sm”) [26].

We also assessed the effect of finding power-law fits to the lemmas of the words and when removing stop words from the corpus. Lemmatization removes the inflectional endings of words and results in a common base word. For example, the lemma of the words “playing,” “plays,” and “played” is “play”. Lemmatization was done using the NLTK Python library WordNet lemmatizer. We used the set of default stop words from the spaCy library.

## RESULTS

### *Power-law Fits*

**Table 1** includes the parameters of the power-law distribution fit to discharge reports and their subsections and the goodness-of-fit test of the power-law distribution. **Table 2** provides the result of the likelihood ratios for the alternative distributions and the p-values for each likelihood test. The last column of **Table 2** lists the statistical support [24] for the power-law hypothesis for each discharge report subsection.

Discharge reports do not follow a pure power-law distribution (P=0.000). The power-law distribution fits the tail of the distribution (word frequency < 1000, see **Fig. 1**), but it does not fit the head of the distribution. Discharge reports are best fit by a truncated power-law (LR=-13.292, P=0.000) and lognormal (LR=-11.805, P=0.000) distributions. The superior fit by the truncated power-law suggests that discharge reports are power-law distributed over a subset of the data as opposed to over the entire data range, and as such are near-Zipfian [12].

**Table 1.** The parameters of the fits of the power-law distribution to the data and the goodness-of-fit test of the power-law distribution for the discharge reports and subsections.

| | Parameters | | | | |
|---|---|---|---|---|---|
| Data set | α | α CI | $x_{min}$ | $x_{min}$ CI | P-value |
| Discharge | 1.500 | (1.497, 1.504) | 3 | (3, 3) | 0.000 |
| Allergies | 1.801 | (1.712, 1.868) | 8 | (2, 16.05) | **0.553** |
| Social history | 1.717 | (1.664, 1.756) | 14 | (4, 27) | **0.665** |
| Past medical history | 1.597 | (1.585, 1.609) | 2 | (2, 5) | 0.000 |
| Family history | 1.653 | (1.627, 1.686) | 2 | (2, 6) | 0.022 |
| History of present illness | 2.072 | (1.510, 2.150) | 854 | (2, 1117.4) | **0.877** |

**Explanatory Notes**: In this table, we present the estimated values for the exponent α, and the minimum value $x_{min}$ for the power-law distribution, and the computed confidence intervals (95%) for these values. In the last column, we present the P-value for the goodness-of-fit test for the power-law distribution. If P-value > 0.10, then the power-law is not ruled out as a plausible distribution for the data. If P-value ≤ 0.10, then we can rule out the power-law distribution as a plausible distribution for the data. Statistically significant values are presented in **bold**.

Allergies (P=0.553), social history (P=0.665), and history of present illness (P=0.877) follow the power-law distribution, but the lognormal and the power-law with cut-off distributions are also plausible fits for the data (see **Table 2**). History of present illness follows a power-law distribution, but the fit comes at the expense of ignoring a large portion of the tail of the data ($x_{min} = 854$). Therefore, the long-tail of the data ($x_{min} < 854$) may best be fit by an alternative distribution. The lognormal and power-law with cut-off distributions are also plausible fits for history of present illness. Past medical history (P=0.000) and family history (P=0.022) do not follow a power-law. Past medical history is best fit by a lognormal, stretched exponential, or a power-law with cut-off. For family history, the best fit is achieved with a power-law with exponential cut-off (LR=-5.495, P=0.001).

**Table 2.** Tests results for power-law distribution and alternative distributions (lognormal, exponential, stretched exponential, and power-law with an exponential cut-off distributions) as a good fit to the data.

| | Power-law | Lognormal | | Exponential | | Stretched exponential | | Power-law with cut-off | | Support for power-law |
|---|---|---|---|---|---|---|---|---|---|---|
| Data set | p | LR | p | LR | p | LR | p | LR | p | |
| Discharge | 0.000 | -11.805 | **0.000** | 38.387 | **0.000** | -0.639 | 0.523 | -13.292 | **0.000** | with cut-off |
| Allergies | **0.553** | 0.688 | 0.491 | 3.938 | **0.000** | 3.551 | **0.000** | 0.293 | 0.376 | moderate |
| Social history | **0.665** | -0.621 | 0.535 | 7.162 | **0.000** | 4.801 | **0.000** | -0.105 | 0.743 | moderate |
| Past medical history | 0.000 | -8.867 | **0.000** | 15.880 | **0.000** | -7.633 | **0.000** | -5.560 | **0.000** | with cut-off |
| Family history | 0.022 | -1.035 | 0.301 | 16.979 | **0.000** | 2.147 | **0.032** | -5.495 | **0.001** | with cut-off |
| History of present illness | **0.877** | -0.284 | 0.776 | 6.240 | **0.000** | 0.774 | 0.439 | -1.219 | 0.320 | moderate |

Explanatory Notes: For each data set we give a P-value for goodness-of-fit test for the power-law distribution. If P-value > 0.10, then the power-law is not ruled out as a plausible distribution for the data. If P-value ≤ 0.10, then we can rule out the power-law distribution as a plausible distribution for the data. We report also the log-likelihood ratios for the comparisons between the power-law distribution hypothesis and the alternative distributions hypotheses, and p-values for the statistical significance of the observed sign (positive or negative) of the log-likelihood ratio for each of the likelihood ratio tests. Positive values of the log-likelihood ratios, with P-values < 0.10, indicate that the power-law distribution is favored over the alternative distribution. Negative values of the log-likelihood ratios, with P-values < 0.10, indicate that the alternative distribution is favored over the power-law distribution. Statistically significant P-values are denoted in **bold**.

The final column of the table lists the statistical support for the power-law hypothesis for each data set as presented in [24]. "Moderate" indicates that the power-law is a good fit but that there are other plausible alternatives as well, and "with cut-off" means that the power-law with exponential cutoff is clearly favored over the pure power-law.

Out of all the distributions tested, the exponential distribution resulted in the worst fits for the discharge reports and all the subsections, followed by the stretched exponential distribution. The exponential distribution has a light-tail, which may explain the poor fit to long-tailed data characteristic of power-laws. **Fig. 1** Illustrates the CCDF of the fit of the power-law and alternative distributions to discharge reports and their subsections. We excluded the exponential fit from **Fig. 1** due to its poor performance, which made it difficult to appreciate the differences in the plot of the various distribution fits.

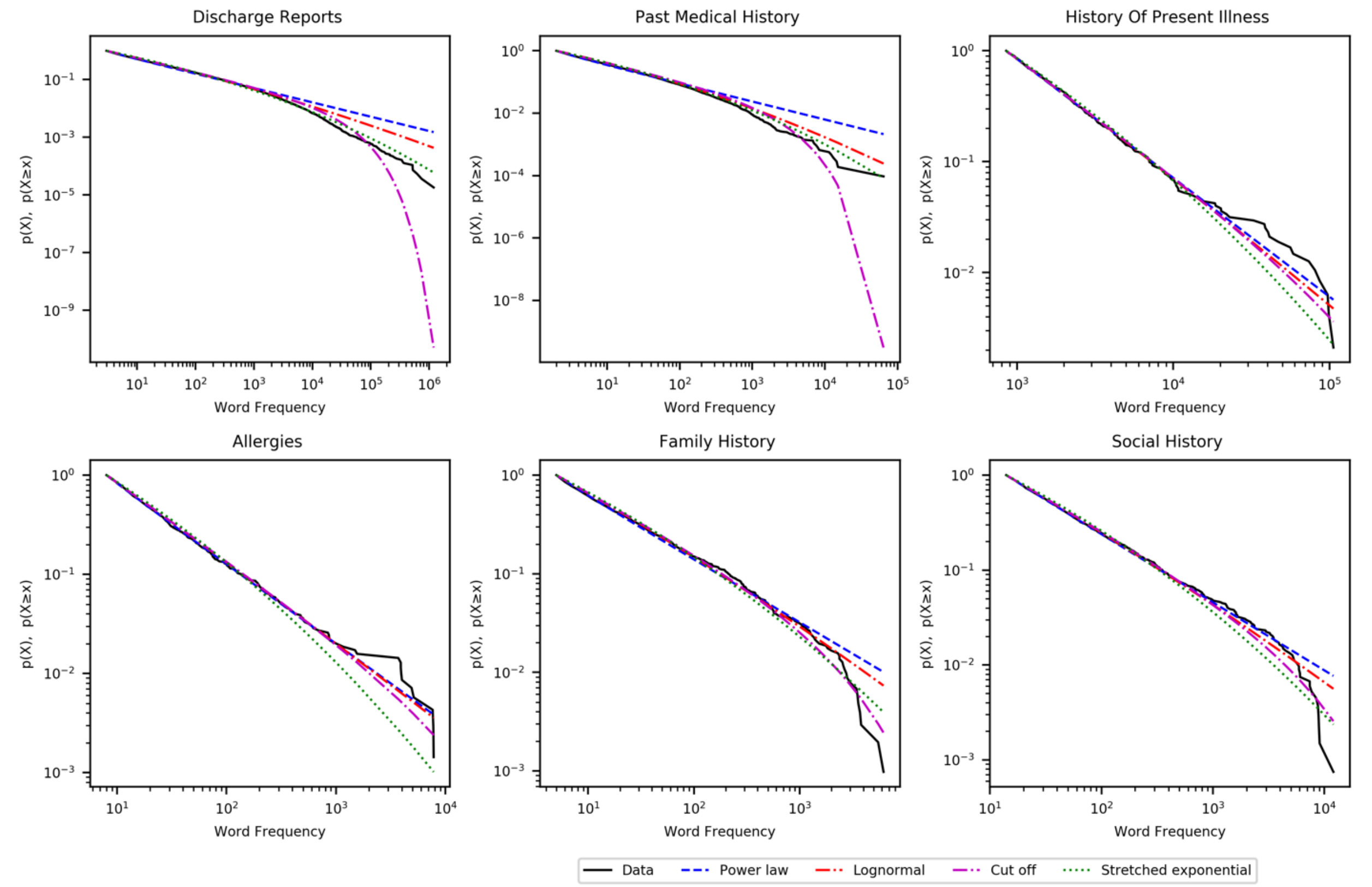

**Figure 1.** CCDF plots showing fits of Zipf's law and lognormal distribution to word frequency for discharge reports, past medical history, history of present illness, allergies, family history, and social history.

The top five words from discharge reports and each subsection are included in **Table 3**, with the top 20 words listed in Appendix C. Numerical tokens ("digit") were part of the top words for discharge reports and all subsections, with the exception of allergies. Digit being a top word is likely due to our normalization which converted all numeric tokens to a single digit token. For allergies, the commonality of the phrase "no known allergies" explains the tokens "no", "known", and "allergies" being in the top five words. For discharge reports and all subsections, stop words and numerical tokens are usually the top words.

***Alternative Tokenization***

The parameters of the power-law fits and the results of the likelihood ratio tests to data tokenized with an alternative tokenizer (spaCy) are included in Appendix D. With alternative tokenization, the results are consistent with our original tokenization (described in prior subsection). Discharge reports do not follow a pure power-law distribution (P=0.000). The truncated power-law (LR=-9.901, P=0.000) and lognormal (LR=-17.787, P=0.000) distributions provide better fits to the data. Allergies (P=0.752), social history (P=0.571), and history of present illness (P=0.488) follow a power-law distribution, and the lognormal and power-law with cut-off distributions also provide suitable fits to the data. The only difference is that family history has a better power-law fit when using the spaCy tokenizer (p=0.102), though the truncated power-law (LR=-5.859, P=0.000) provided the best fit to family history. Across all of the subsections, the truncated power-law provided the best fits.

**Table 3.** Top five words from discharge reports and each subsection

| Dataset | Top words | Top words with lemmatization and stop words removed |
|---|---|---|
| Discharge reports | digit, the, and, was, of | digit, mg, patient, tablet, po |
| Allergies | allergies, no, known, patient, to | allergy, drug, known, patient, recorded |
| Family history | of, digit, died, father, with | digit, died, father, mother, cancer |
| History of present illness | digit, and, the, was, to | digit, patient, pain, day, history |
| Past medical history | digit, of, in, and, sp | digit, sp, disease, history, hypertension |
| Social history | digit, a, in, and, lives | digit, life, year, use, alcohol |

***Lemma Forms***

The parameters of the power-law fits to data when removing stop words and using the lemmas of the discharge reports and the results of the likelihood ratio tests are included in Appendix E. Discharge reports do not follow a pure power-law when removing stop words and using the lemmas of the discharge reports. The truncated power-law (LR=-10.661, P=0.000) and lognormal (LR=-12.454, P=0.000) distributions provided superior fits to discharge reports. These results are consistent with the word forms of discharge reports. For each of the subsections, lemmatization and removing stop words made the truncated power-law provide the best fit across all the subsections, with the lognormal distribution being the second best fit. The consistency of the parameter fits found for word forms vs lemmas is consistent with prior work [22]. The top words with lemmatization and stop words removed are listed in **Table 3**.

## DISCUSSION

### *Main Findings*

Discharge reports appear to be near-Zipfian by having the truncated power-law consistently providing superior fits over a pure power-law. Discharge report subsections have some support for the pure power-law, with the support being either moderate or truncated. The lognormal distribution also provided strong fits to discharge reports and subsections. Past medical history was the only subsection that had no appropriate fit amongst the distributions tested.

The superior fit of the truncated power-law over the pure power-law suggests word frequency in medical discharge reports follows a power-law over some range of the data. Specifically, it was the tokens with the highest frequencies which deviated from the pure power-law. This may be due to the way discharge reports are written, large medical vocabulary, and commonly used medical abbreviations to describe prescriptions, measurements, symptoms, and test results. Zipf's law is often observed when words are ambiguous and have multiple meanings, and generic words are indeed more frequent than specific words. Describing medical conditions requires specificity in discharge summaries, which may explain generic terms appearing less frequently and causing a weaker fit to Zipf's distribution, although confirmation of this hypothesis remains to be verified.

We found alternative tokenizations to yield consistent results. The different tokenizations are more likely to affect the tail of the distribution, especially the tokens with counts equal to one. For instance, depending on the tokenization of numbers, this can result in a large number of numeric tokens with a count of one (i.e. "2.8", "110/65", "77mm").

Lemmatization and removing stop words yielded consistent results when using word forms and keeping stop words in the text. Lemmatization, stemming, and removing stop words are common pre-processing procedures in some natural language processing tasks. Thus, our results give confidence that whether or not these pre-processing steps are applied, researchers can expect consistency in (1) the parameters of the power-law fits to the data, and (2) the selection of alternative distributions that provide superior fits to the power-law. Our results also support prior work which showed consistent parameter fits for word forms vs lemmas [22].

Numerical tokens had the highest word frequency in our results. This reflects our parsing approach, which grouped all numerical tokens, suggesting the need for a more detailed numerical parsing and extraction given that numbers in discharge reports contain critical information such as results from pathological tests, vital signs, medication dosages, and physiological measurements. The top words from the discharge report subsections indicate that these sections contain language that is not specific to ICU conditions, and include wording commonly used for documenting patient history during GP consultations as part of the generally accepted structured SOAP summary [27].

Prior work has argued that Zipf’s law theoretically arises from the competing pressures of minimizing the effort to communicate by a speaker and a listener [28]. We speculate that clinical documentation, including the preparation of discharge reports, operates in an environment that gives rise to competing pressures of accurate communication versus efficiency. The competing pressures are the documentation burden of generating an accurate and detailed discharge report vs the need for the information in the discharge report to be understood by the patient, the primary care clinician, and even the same hospital staff if the patient is readmitted for hospitalization.

**Implications for Machine Learning**

Our work has implications for parametric and non-parametric Bayesian modelling [10,29] and methods bridging Bayesian non-parametrics with Bayesian deep learning [30]. Our results support truncated power-law and the lognormal distributions as appropriate priors for modelling medical discharge report word frequency. Models have been tailored with a power-law prior for language modelling [14], topic models [15,16], term frequency [31], and word embeddings [32]. For example, a topic model such as Latent Dirichlet Allocation (which relies on the Dirichlet distribution to model word distribution) cannot capture the power-law behaviour that arises in corpora—resulting in less descriptive topics and uninformative results [15,33,34]—whereas using models that capture the power-law behaviour of a dataset can improve learning [35,36].

In probabilistic programming, practitioners have the option of explicitly coding power-law and lognormal priors. In other cases, Bayesian nonparametric models for modeling power-law behavior can be used, such as the Pitman-Yor process [37], the Chinese restaurant process [38], and the Indian Buffet Process [36]. The Pitman-Yor process has had strong applications for text applications, including language models and topic modeling [14,15,39,40]. The parameters and hyperparameters of Bayesian non-parametric models may be used as a proxy for the exponential cut-off, such as the discount parameter of the Pitman-Yor process used to control the tail behavior.

Our findings showing that discharge reports are near-Zipfian suggest that accounting for any power-law behavior is likely to benefit modeling approaches (even without the cut-off). Prior work in image segmentation combining hierarchical Pitman-Yor processes with Gaussian processes [41] also show the potential for compositional combination of probabilistic models for producing power-law priors. Finally, we hypothesize that the deviation from the pure power-law in the head of the data (strong Zipfian fit in the tail and a weaker fit in the head) may be explained by a double power-law behavior [42,43], with models capturing this behavior (the generalized BRFY process [43] and the Beta prime process [44,45]) likely improving modeling efficacy.

## Comparison with Prior Work

This study is the first to demonstrate that medical discharge reports are near-Zipfian and are better fit by a truncated power-law. Specifically, the analysis focused on full written text in electronic health records, as opposed to examining medical phrases [18] or medical codes [17–19]. A prior study showed CCDF plots of the Read codes for diagnoses and procedures where the heads of the data deviated considerably from Zipf's law [17]. Our CCDF plots show a similar behaviour, with the power-law fitting the tail of the distribution properly, and deviating from the head of the discharge report data and subsections. Two other studies [18,19] also showed that the terms with the highest frequency (rank 1-10) deviated from the power-law, and the long-tail was fit well by the power-law distribution.

## Limitations

Limitations of our work include our parsing approach, which resulted in the tokens used for calculating word frequency and distribution fits. Future should address how modelling of numerical tokens in discharge reports affect the power-law distribution fits. Empirical evidence in some datasets has shown that these may have a double power-law fit, where high-frequency words follow a power-law distribution with one set of parameters, and low-frequency words follow a power-law distribution with a different set of parameters [42,43]. Future work should explore optimal ways of finding piece-wise power-law fits to medical discharge reports.

The MIMIC dataset is comprised of patients admitted to critical care units at a single hospital [21]. As such, the discharge reports of ICU patients may not be an appropriate representation of discharge reports of all hospitalized patients. Further work is also needed to determine the generalizability of our findings with discharge reports generated from hospitals with different characteristics, such as teaching status, urban location, geographic region, and bed size.

## CONCLUSION

Discharge reports are near-Zipfian by following a truncated power-law. The data was also fit well by the lognormal distribution. AI and ML modelling approaches to discharge reports can benefit from handling the truncated power-law properties of discharge reports and their subsections. Our work presents evidence for using a truncated power-law prior for Bayesian modelling of medical text.

**Acknowledgments**

This research was supported by the National Health and Medical Research Council (NHMRC) grant APP1134919 (Centre for Research Excellence in Digital Health).

**Author Contributions**

J.C.Q. designed the project with input from L.L., A.B.K., S.B., F.R. and E.C. J.C.Q performed the computational work. J.C.Q. and C.T. analysed the results. J.C.Q. wrote the first draft and prepared the tables and the figures. J.C.Q. and C.T. wrote the Methods. C.T. wrote the explanatory notes for the tables. L.L., C.T., A.B.K., S.B., F.R. and E.C. critically reviewed and revised the writing of the final text. All authors approved the final draft.

**Competing Interests**

None.

## APPENDICES

### APPENDIX A: PARSING DETAILS

The MIMIC-III dataset analysed in the present study is freely and publicly available at https://mimic.physionet.org/. We normalized the text by applying lowercasing and converting all digits to a digit token "D" [1]. All numeric tokens with a period, slash, or hyphen, were replaced with a single token ("D.D", "D/D", "D-D", "D:D" all converted to token "digit"). We split the text into tokens by removing all punctuation symbols, removing all text associated with de-identified data (e.g. dummy clinician names, dummy dates, dummy hospital names), and splitting using whitespace as the delimiter. We excluded all words with a frequency of one as some of these may be the result of typos or parsing errors [2].

### APPENDIX B: TESTING POWER-LAW HYPOTHESIS

First, the power law hypothesis was tested. That is, given the MIMIC-III data set and the statistical hypothesis that the data from the empirical real data are drawn from a power law distribution, we examined the goodness-of-fit test, using a bootstrapping hypothesis test, to verify if the statistical hypothesis was plausible. A P-value was computed, quantifying the plausibility of the statistical hypothesis. P-values > 0.10 indicate that the statistical hypothesis is a plausible one, that is, that the power law is not ruled out as a plausible distribution for the data [1]. The power law is ruled out as a plausible distribution for the data if P-value ≤ 0.10 [1].

Second, we explored if alternative distributions might yield a fit as good as or better than the power-law distribution using a likelihood ratio test (LR) [1]. In our case, the likelihood ratio test computes the likelihood of the data under two competing distributions for the data specified by two competing statistical hypotheses (the power-law distribution hypothesis vs the alternative distribution hypothesis). The alternative hypotheses (Table 1) included the lognormal, exponential, stretched exponential, and truncated power-law (also referred to as power-law with an exponential cut-off), with the choice of these alternative hypotheses based on prior work [1]. The logarithm of the ratio of the two likelihoods of the data under two competing distributions specified by two competing statistical hypotheses indicates which distribution is a better fit for the data. The log-likelihood ratio may be positive or negative, depending on which distribution is a better fit for the data. A log-likelihood ratio of 0 indicates that one distribution compared to the other distribution is not better fit to the data. Positive log-likelihood ratios that are statistically significant (P-value < 0.10) indicate that the power-law distribution is favoured over the alternative distribution. Negative log-likelihood ratios that are statistically significant (P-value < 0.10) indicate that the alternative distribution is favoured over the power-law distribution. A P-value ≥ 0.10 indicates that the test does not favour the power-law distribution over the alternative distribution (no statistical significance).

**Table B.1.** Alternative distributions compared to the power-law distribution.

| Name | Distribution |
|---|---|
| Power-law with cut-off | $f(x) \propto x^{-\alpha} e^{-\lambda x}$ |
| Exponential | $f(x) \propto e^{-\lambda x}$ |
| Stretched exponential | $f(x) \propto x^{-\alpha} e^{-\lambda x}$ |
| Lognormal | $f(x) \propto \frac{1}{x} \exp\left[-\frac{(\ln x - u)^2}{2\sigma^2}\right]$ |

## APPENDIX C: TOP 20 WORDS FOR DISCHARGE REPORTS AND SUBSECTIONS

| **Discharge reports** | | **Family history** | | **History of present illness** | |
|---|---|---|---|---|---|
| digit | 1193115 | of | 6211 | digit | 106119 |
| the | 740201 | digit | 5521 | and | 99166 |
| and | 618235 | died | 3811 | the | 97442 |
| was | 525370 | father | 3722 | was | 86696 |
| of | 520107 | with | 3580 | to | 80825 |
| to | 518540 | mother | 3523 | of | 73819 |
| with | 367382 | and | 3447 | with | 58714 |
| a | 347903 | at | 3010 | a | 56266 |
| on | 338804 | cancer | 2815 | he | 47004 |
| in | 277326 | no | 2761 | on | 41345 |
| for | 253278 | history | 2654 | in | 40489 |
| mg | 238542 | in | 2581 | for | 38881 |
| no | 212111 | noncontributory | 2533 | she | 38374 |
| patient | 211207 | disease | 2379 | patient | 32747 |
| is | 179824 | age | 2375 | at | 22964 |
| he | 175136 | family | 2296 | had | 21911 |
| po | 169243 | had | 1990 | is | 20328 |
| tablet | 154316 | a | 1829 | his | 20198 |
| at | 151326 | mi | 1784 | pain | 18716 |
| blood | 142950 | brother | 1529 | her | 18544 |

| **Past medical history** | | **Social history** | | **Allergies** | |
|---|---|---|---|---|---|
| digit | 64336 | digit | 12043 | allergies | 7826 |
| of | 15146 | a | 9061 | no | 7774 |
| in | 14516 | in | 8932 | known | 7668 |
| and | 13050 | and | 8849 | patient | 5147 |
| sp | 12214 | lives | 8746 | to | 4912 |
| with | 11398 | with | 8431 | as | 4014 |
| on | 8381 | he | 7684 | drugs | 3953 |
| to | 8092 | is | 7395 | recorded | 3918 |
| disease | 7832 | the | 7389 | having | 3913 |
| history | 7558 | no | 6049 | drug | 3717 |
| hypertension | 7227 | use | 6032 | penicillins | 1562 |
| post | 6992 | she | 5912 | the | 1496 |
| status | 6915 | of | 5909 | and | 1161 |
| the | 6847 | has | 5659 | has | 1003 |
| a | 4965 | alcohol | 5336 | sulfa | 885 |
| for | 4936 | tobacco | 5266 | codeine | 867 |
| left | 4833 | years | 5155 | reactions | 863 |
| chronic | 4629 | patient | 4829 | adverse | 845 |
| artery | 4160 | to | 4477 | penicillin | 689 |
| right | 3976 | history | 4257 | iodine | 621 |

**APPENDIX D: PARAMETER FITS AND ALTERNATIVE DISTRIBUTION RESULTS USING AN ALTERNATIVE TOKENIZER**

Table D.1. The parameters of the fits of the power-law distribution to the data and the goodness-of-fit test of the power-law distribution for the discharge reports and subsections tokenized with spaCy.

| | Parameters | | | | |
|---|---|---|---|---|---|
| | α | α CI | $x_{min}$ | $x_{min}$ CI | p-value |
| Discharge | 1.506 | (1.503, 1.509) | 2 | (2, 2) | 0.000 |
| Allergies | 1.819 | (1.707, 1.885) | 9 | (2, 16.05) | **0.752** |
| Social history | 1.702 | (1.663, 1.737) | 8 | (5, 21) | **0.571** |
| Past medical history | 1.591 | (1.580, 1.603) | 2 | (2, 5) | 0.000 |
| Family history | 1.636 | (1.608, 1.666) | 2 | (2, 6) | **0.102** |
| History of present illness | 2.100 | (1.497, 2.207) | 941 | (2, 1477.1) | **0.488** |

Table D.2. Tests results for power-law distribution and alternative distributions (lognormal, exponential, stretched exponential, and power-law with an exponential cut-off distributions) as a good fit to the data tokenized with spaCy.

| | | Lognormal | | Exponential | | Stretched exponential | | Power-law with cut-off | | Support for power-law |
|---|---|---|---|---|---|---|---|---|---|---|
| Data set | p | LR | p | LR | p | LR | p | LR | p | |
| Discharge | 0.000 | -17.787 | **0.000** | 51.881 | **0.000** | 51.044 | **0.000** | -9.901 | **0.000** | with cut-off |
| Allergies | **0.752** | 0.121 | 0.904 | 8.424 | **0.000** | 1.820 | 0.069 | -1.563 | 0.367 | moderate |
| Social history | **0.571** | -1.039 | 0.299 | 15.310 | **0.000** | 1.739 | 0.082 | -4.023 | **0.001** | with cut-off |
| Past medical history | 0.000 | -8.876 | **0.000** | 16.269 | **0.000** | -7.939 | **0.000** | -5.592 | **0.000** | with cut-off |
| Family history | **0.102** | -1.738 | 0.082 | 16.815 | **0.000** | 1.714 | 0.086 | -5.859 | **0.000** | with cut-off |
| History of present illness | **0.488** | 0.090 | 0.928 | 6.322 | **0.000** | 1.283 | 0.199 | -1.030 | 0.442 | moderate |

**APPENDIX E: PARAMETER FITS AND ALTERNATIVE DISTRIBUTION RESULTS FOR LEMMAS**

Table E.1. The parameters of the fits of the power-law distribution to the data and the goodness-of-fit test of the power-law distribution for the discharge reports and subsections with lemmatization and without stop words.

| | Parameters | | | | |
|---|---|---|---|---|---|
| | α | α CI | $x_{min}$ | $x_{min}$ CI | p-value |
| Discharge | 1.511 | (1.508, 1.515) | 3 | (3, 3) | 0.000 |
| Allergies | 1.881 | (1.797, 1.959) | 9 | (5, 18) | **0.631** |
| Social history | 1.815 | (1.761, 1.868) | 14 | (8, 31) | **0.961** |
| Past medical history | 1.606 | (1.593, 1.618) | 2 | (2, 5) | 0.000 |
| Family history | 1.699 | (1.668, 1.738) | 2 | (2, 5) | **0.536** |
| History of present illness | 1.525 | (1.518, 1.532) | 2 | (2, 3) | 0.000 |

Table E.2. Tests results for power-law distribution and alternative distributions (lognormal, exponential, stretched exponential, and power-law with an exponential cut-off distributions) as a good fit to the data with lemmatization and without stop words.

| | | Lognormal | | Exponential | | Stretched exponential | | Power-law with cut-off | | Support for power-law |
|---|---|---|---|---|---|---|---|---|---|---|
| Data set | p | LR | p | LR | p | LR | p | LR | p | |
| Discharge | 0.000 | -12.454 | **0.000** | 51.986 | **0.000** | -1.343 | 0.179 | -10.661 | **0.000** | with cut-off |
| Allergies | **0.631** | -0.050 | 0.960 | 6.106 | **0.000** | 1.208 | 0.227 | -1.024 | 0.581 | moderate |
| Social history | **0.961** | -0.743 | 0.457 | 9.938 | **0.000** | 0.740 | 0.459 | -2.495 | **0.026** | with cut-off |
| Past medical history | 0.000 | -8.139 | **0.000** | 12.424 | **0.000** | -5.864 | **0.000** | -8.993 | **0.000** | with cut-off |
| Family history | **0.536** | -1.019 | 0.308 | 10.985 | **0.000** | 0.966 | 0.334 | -3.211 | **0.005** | with cut-off |
| History of present illness | 0.000 | -12.875 | **0.000** | 27.300 | **0.000** | -12.274 | **0.000** | -9.759 | **0.000** | with cut-off |